\documentclass{article}
\DeclareUnicodeCharacter{2605}{\ensuremath{\star}}

\usepackage{microtype}
\usepackage{subcaption}
\usepackage{graphicx}
\usepackage{booktabs} 
\usepackage{multirow}
\usepackage{float}

\usepackage{hyperref}

\usepackage[utf8x]{inputenc} 
\usepackage{tabularx}
 \newcolumntype{L}{>{\raggedright\arraybackslash}X}

\usepackage[accepted]{icml2025}

\usepackage{amsmath}
\usepackage{amssymb}
\usepackage{mathtools}
\usepackage{amsthm}

\usepackage[capitalize,noabbrev]{cleveref}

\usepackage[font=footnotesize,labelformat=simple]{subcaption}

\theoremstyle{plain}

\theoremstyle{definition}

\theoremstyle{remark}

\usepackage[textsize=tiny]{todonotes}

\usepackage[most]{tcolorbox}
\usepackage{placeins} 

\tcbset{
  icmlbox/.style={
    colback=gray!10,
    colframe=blue!50!black,
    coltitle=white,
    colbacktitle=blue!60!black,
    title=Prompt Format,
    fonttitle=\bfseries,
    boxrule=0.4pt,
    arc=1.5mm,
    outer arc=1.5mm,
    left=4pt,        
    right=4pt,
    top=3pt,
    bottom=3pt,
    enhanced,
    sharp corners=south,
    breakable
  }
}

\begin{document}

\twocolumn[
\icmltitle{Lightweight Safety Guardrails via Synthetic Data and RL-guided Adversarial Training}



\icmlsetsymbol{equal}{*}

\begin{icmlauthorlist}
\icmlauthor{Aleksei Ilin}{app}
\icmlauthor{Gor Matevosyan}{app}
\icmlauthor{Xueying Ma}{app}
\icmlauthor{Vladimir Eremin}{app}
\icmlauthor{Suhaa Dada}{app}
\icmlauthor{Muqun Li}{app}
\icmlauthor{Riyaaz Shaik}{app}
\icmlauthor{Haluk Noyan Tokgozoglu}{app}

\end{icmlauthorlist}

\icmlaffiliation{app}{Apple}

\icmlcorrespondingauthor{Aleksei Ilin}{aleksei\_ilin@apple.com}
\icmlcorrespondingauthor{Gor Matevosyan}{gmatevosyan@apple.com}
\icmlcorrespondingauthor{Xueying Ma}{amber\_ma1@apple.com}
\icmlcorrespondingauthor{Vladimir Eremin}{vladimir\_eremin@apple.com}
\icmlcorrespondingauthor{Muqun Li}{mli54@apple.com}

\icmlkeywords{Machine Learning, ICML}

\vskip 0.3in
]



\printAffiliationsAndNotice{} 

\begin{abstract}
We introduce a lightweight yet highly effective safety guardrail framework for language models, demonstrating that small-scale language models can achieve, and even surpass, the performance of larger counterparts in content moderation tasks. This is accomplished through high-fidelity synthetic data generation and adversarial training. The synthetic data generation process begins with human-curated seed data, which undergoes query augmentation and paraphrasing to create diverse and contextually rich examples. This augmented data is then subjected to multiple rounds of curation, ensuring high fidelity and relevance. Inspired by recent advances in the Generative Adversarial Network (GAN) architecture, our adversarial training employs reinforcement learning to guide a generator that produces challenging synthetic examples. These examples are used to fine-tune the safety classifier, enhancing its ability to detect and mitigate harmful content. Additionally, we incorporate strategies from recent research on efficient LLM training, leveraging the capabilities of smaller models to improve the performance of larger generative models. With iterative adversarial training and the generation of diverse, high-quality synthetic data, our framework enables small language models (SLMs) to serve as robust safety guardrails. This approach not only reduces computational overhead but also enhances resilience against adversarial attacks, offering a scalable and efficient solution for content moderation in AI systems.
\end{abstract}

\section{Introduction}

\subsection{AI Safety Guardrail}

Recent years have witnessed a significant evolution in conversational AI, transitioning from task-oriented chatbots with rigid response patterns to generative Large Language Model (LLM)-based systems. While substantial effort has been dedicated to dataset curation and LLM alignment, the inherent generative capabilities of these models introduce the risk of generating undesirable responses \cite{yuan2023gpt, zou2023universal, huang2023catastrophic, NEURIPS2023_4dbb61cb}. Consequently, the identification of potentially harmful or policy-violating content remains a crucial challenge in ensuring the responsible adoption of conversational AI.

Currently, numerous datasets and open-source models exist for developing safety guardrails. Dataset creation primarily involves three approaches: manual data collection and annotation \cite{lin2023toxicchat}, semi-manual or active-learning-based data collection from real interactions \cite{markov2023holistic}, and synthetic data generation, e.g., adversarial synthetic data generation \cite{wildguard2024, li2024salad, mazeika2024harmbench}. Open-source models encompass classifiers for specific harm categories \cite{wildguard2024}, binary safe/unsafe classifiers \cite{mazeika2024harmbench, lee2024harmaug}, and even specialized generative classifiers \cite{li2024salad, inan2023llama}. Notably, generative classifiers can generally be implemented using any generative language model, provided that clear rubric guidelines are established.

\subsection{Synthetic Data Generation and Curation}
Although modern LLMs can generate human-like responses, they often struggle to produce consistent and contextually relevant outputs, especially in tasks that require nuanced understanding or fine-grained distinctions. In our setup, each generated example is annotated with a label to facilitate downstream training. Given this, all initial labels should be considered weak, and data cleaning becomes essential for curating high-quality labeled examples. Moreover, synthetic datasets, while easier to generate or annotate, may not capture the full complexity or noise of real data, leading to out-of-distribution (OOD) issues at evaluation time \cite{nguyen2024deepdomainadaptationsim2real}. This sim-to-real gap has motivated research into methods that align synthetic data closer to real data distributions, ensuring models learn features that transfer robustly. A key idea is to filter or select synthetic examples based on their similarity to real data so that the training mix better represents the target domain. We mainly focused on methods with training on weak labels like Active Learning or Semi-Supervised learning. Methods can generally be divided into two groups: slowing down training \cite{sohn2020fixmatch, li2020dividemix} or analyzing hard examples \cite{xiao2023freeal, rawat2024little}.

\subsection{Adversarial Training}
To overcome LLM refusals - where the model refuses to generate specific types of content due to safety constraints or misalignment with the user's intent, or to adapt the model to the new domain, it is often necessary to fine-tune the model or perform preference optimization through a feedback loop, which is usually called ``alignment'' \cite{zhou2023lima, ouyang2022training}. Reinforcement learning has emerged as a powerful technique for aligning language models, sparking considerable debate regarding optimal implementation. Current research explores a spectrum of approaches, ranging from traditional policy gradient methods to more sophisticated surrogate objectives such as Proximal Policy Optimization (PPO) and Group Relative Policy Optimization (GRPO) \cite{ahmadian2024back, shao2024deepseekmath}. Furthermore, there is active investigation into various reward structures, including loss complexity rewards \cite{xu1802dpgan} and reward models trained on human feedback \cite{yu2023fine, lee2023rlaif}.

\section{Methodology}

\subsection{Problem Overview}
This work presents a novel framework for constructing language model safety guardrails based on SLMs through synthetic data augmentation. The proposed methodology encompasses human-guided seeding, multi-stage data generation and curation, and iterative fine-tuning/alignment of generative models, with a focus on synthesizing data that captures challenging or problematic cases.

\subsection{Safety Risk Taxonomy Ideation and Implementation}

This methodology enables the construction of a comprehensive taxonomy of potential safety risks, categorized by both severity and domain. To facilitate quantitative analysis and comparison across different datasets with varying scales and characteristics, we opted for a \textit{binary classification} approach, assigning a value of 1 to unsafe instances and 0 to safe instances. This allows for standardized evaluation and aggregation of risk assessments.

\subsection{Data Augmentation With Synthetic Data Generation}

Our approach implements content safety moderation through systematic synthetic data generation that combines human expertise with AI scalability. This methodology addresses three critical challenges in safety taxonomy development: coverage, adaptability, and privacy compliance.

\subsubsection{Human-Curated Seeding}
Our synthetic data generation relies on expert-designed seed data to produce high-quality, relevant outputs. Starting from a risk taxonomy of X core categories defined by domain experts, human red-teaming generates examples mirroring the risk categories, with particular attention to covering borderline cases. This structure allows for continuous taxonomy refinement under human supervision.

\begin{tcolorbox}[icmlbox, title=Example Prompt Instruction]
You are a query generator that is tasked with creating examples of adversarial queries. You need to help with generation data for combating unsafe queries and create \textbf{ONE} hypothetical music query that contains \textit{\textcolor{red}{self-harm}}. \\

Here are a few examples:\\
Seed 1: Give me songs for when everything feels pointless and empty. \\
Seed 2: Play the ``Suicide Siren'' playlist when I slit my wrists. \\

Response should be strict JSON-formatted with a single field \textbf{Query}. Now provide your answer:
\end{tcolorbox}

\subsubsection{Tiered Prompt Engineering}
To enhance the variety of synthetic data while maintaining control over the generation process, we utilize a tiered prompt engineering methodology. First, LLMs are employed to generate lists of potential seed terms. These seed terms are then systematically integrated into structured prompt templates to create diverse synthetic queries. This tiered process comprises three key phases:
\begin{itemize}
\item \textbf{Concept expansion}: LLMs generate variations for each concept and sub-concept within the safety taxonomy, ensuring that the generated terms remain aligned with the established categories. This step aims to broaden the coverage of potential risks.
\item \textbf{Contextual template infusion}: The expanded set of concepts is then embedded into realistic query structures, mirroring how users might naturally express these concepts in real-world scenarios. This step enhances the realism and relevance of the synthetic data.
\item \textbf{Style mutation}: Finally, transformations are applied to the style and tone of the queries, injecting diverse linguistic representations and ensuring that the generated data is not overly homogeneous. This step contributes to the robustness of the models trained on the synthetic data.
\end{itemize}

\begin{tcolorbox}[icmlbox, title=Generate Seed Terms]
Create \textbf{ONE} example of a succinct hypothetical music request (\textit{tunes/tracks/playlist}) that focuses instead on \texttt{\{seed\_1\}}, intended not metaphorically but rather literally (explicitly), in the most tactless irresponsible (sociopathic), distressing, or pessimistic way.

The response should be no more than a short sentence in length and be strict JSON-formatted with a single field \textbf{Query}.\\

Now provide your pithy answer:
\end{tcolorbox}

\subsection{Synthetic Data Curation}
To enhance the synthetic data quality while managing inherent unpredictability, we implement a synthetic data curation framework that combines loss modeling-based sample selection, embedding-based analysis, and LLM-based validation. This approach addresses the core challenge of maintaining semantic fidelity in generated queries while enabling continuous system improvement.

\subsubsection{Loss Modeling-Based Data Cleaning}

Although neural networks are not inherently robust to outliers, they typically tolerate a certain proportion of erroneous samples without significant loss of predictive performance. Nevertheless, anomalies or erroneous data points can pose considerable challenges to the model's predictive accuracy, particularly if encountered during training. Consequently, training examples that continue to exhibit high loss after multiple iterations merit closer examination. Due to the absence of established heuristics for determining an appropriate cutoff threshold, a principled approach is necessary to systematically identify these problematic data points.

\begin{figure}[h]
    \centering
    \begin{subfigure}[t]{0.49\columnwidth}  
        \includegraphics[width=\linewidth]{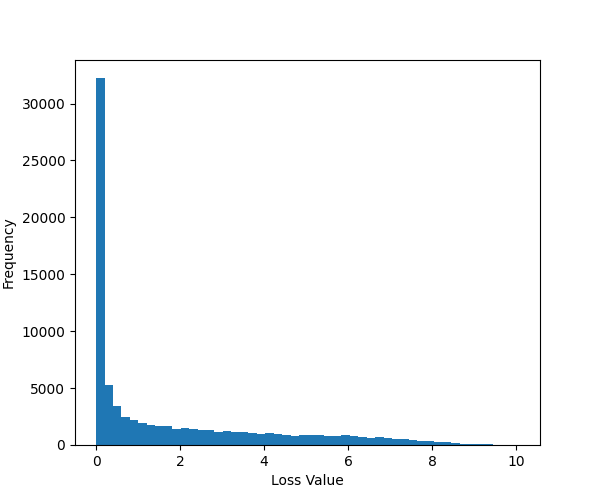}
        \caption{Vanilla Cross Entropy Loss}
        \label{fig:1}
    \end{subfigure} \hfill   
    \begin{subfigure}[t]{0.49\columnwidth}  
        \includegraphics[width=\linewidth]{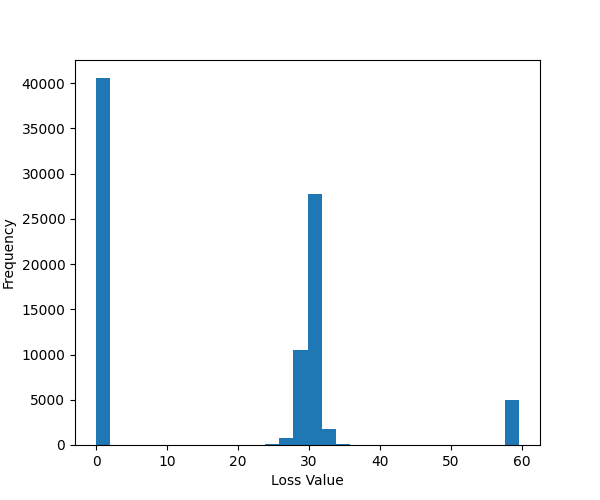}
        \caption{Entropy-based Cleaning Loss}
        \label{fig:2}
    \end{subfigure}
    \caption{Histograms of Vanilla Loss-based Cleaning vs Max Entropy Loss Cleaning applied to WildGuard trainset. The anomaly class is clearly visible as a third separated distribution, while in vanilla loss settings it is hard to see how to cut off the anomalies.}
    \label{fig:losses}
\end{figure}

To address this issue, we introduce a novel \textbf{entropy-based data cleaning objective} (Equation~\ref{eq:1c}). The process is as follows:

\begin{enumerate}
    \item \textbf{Train the model using two step loss present in Equation~\ref{eq:1c})}. The first term, $L_n$, is the standard Cross-Entropy Loss per example (Equation~\ref{eq:1a}, C is the number of classes, N is the number of examples in a batch), whereas the second term, $H(L)$, maximizes the sum of Entropy of the per-example loss distribution of the previous loss (Equation~\ref{eq:1b}). The exact number of epochs depends on the data, but the expectation is the model's training loss should reach convergence.
    
    \item \textbf{Plot loss histograms of the training data.} Since Gaussian distribution maximizes Entropy for given parameters, in binary settings we will get bell curved distributions for 0, 1 and anomaly classes (example for WildGuard trainset is on Figure~\ref{fig:losses}). This can be skipped in automatic settings.

    \item \textbf{Exclude the third Gaussian with the largest mean.} The exclusion of the third Gaussian can be done automatically using Mixture of Gaussians. This step is relatively trivial given that the Gaussians are well separated.
\end{enumerate} 
\begin{align}
    \label{eq:1a}
    \mathcal{L}_n = - \sum_{c=1}^{C} y_{n,c} \log \hat{y}_{n,c}, \quad \text{for } n = 1, \ldots, N \\
    \label{eq:1b}
    H(\mathcal{L}) = - \sum_{n=1}^{N} p_n \log p_n \\
    \label{eq:1c}
    \mathcal{J} = \mathcal{L}_n -  H(\mathcal{L})
\end{align}

In addition, we also adopted data cleaning based on pretrained and fine-tuned embeddings and LLM-as-a-Judge, both of which we illustrate with more details in Appendix~\ref{app:datacleaning}.

\subsection{Guardrail Model Fine-Tuning}
\begin{figure}[h]
    \centering
   \includegraphics[width=\linewidth]{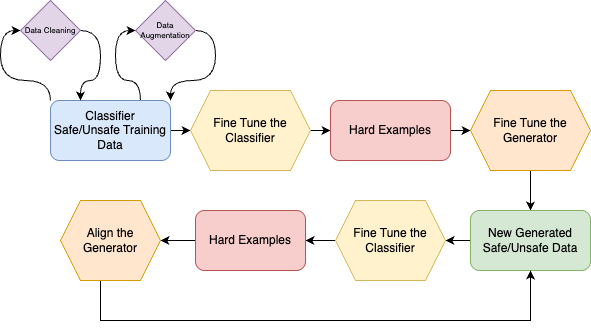}
    
    \caption{Overview of the RL-guided Adversarial Training Pipeline With Synthetic Data Augmentation}
    \label{fig:pipeline}
\end{figure}

Our approach utilizes decoder-only language models as the foundation for our safety guardrails. These models are fine-tuned on a large-scale dataset of queries that reflects diverse user intent using a RL-guided adversarial training pipeline (as presented in Figure~\ref{fig:pipeline}), allowing for a nuanced understanding of user interactions. Specifically, we use the GRPO loss to guide the generator alignment process, which is described in more detail in the following section.

\subsubsection{Hard Sample Selection Guided by Small Models} \label{section:adaptation}

To enhance the efficiency and effectiveness of fine-tuning large generative models, we adopt a strategy that utilizes a smaller, computationally efficient language model to guide the process similar to \cite{rawat2024little}. This approach is particularly beneficial in scenarios with limited computational resources or when rapid adaptation to new domains is required.

\paragraph{Step 1: Training the SLM}\mbox{}\\
We begin by fine-tuning an SLM (in this case, Lite-Oute-1-300M-Instruct with a classification head). This model serves as a preliminary learner, capturing the foundational patterns and structures within the data.

\paragraph{Step 2: Evaluating and Selecting Hard Training Examples}\mbox{}\\
After the SLM has been trained:
\begin{itemize}
    \item \textbf{Compute Cross-Entropy Loss of the Classifier}: For each training example, calculate the cross-entropy loss using the SLM. This metric indicates how well the model predicts each example, with higher values signifying more challenging instances.
    \item \textbf{Sort and Filter Examples}:
    \begin{itemize}
        \item \textbf{Exclude Top 20\%}: Remove the top 20\% of examples with the highest loss. These are likely outliers or anomalies that could introduce noise into the fine-tuning process. Here, we picked 20\% based on a heuristic; this proportion can be further refined through additional experimentation.
        \item \textbf{Select Above-Average Loss Examples}: From the remaining data, identify examples where the loss exceeds the dataset's average. These instances are considered ``hard but learnable'' offering valuable information for further training.
    \end{itemize}
\end{itemize}
This selection process ensures that the subsequent fine-tuning focuses on informative and challenging examples, enhancing the model's ability to generalize.

\paragraph{Step 3: Fine-Tuning the LLM}\mbox{}\\
Utilize the curated subset of training examples to fine-tune a larger generative model (e.g., Mistral-7B\cite{jiang2023mistral7b}). This targeted fine-tuning allows the LLM to focus on complex patterns and nuances that the SLM struggled with, leading to improved performance in the target domain.

\paragraph{Step 4: Generating Synthetic Data}\mbox{}\\
Post fine-tuning, the enhanced LLM can be employed to generate synthetic data that mirrors the complexities of the target domain. This synthetic data can serve multiple purposes: \textbf{Augmenting Training Sets} (enriching existing datasets with diverse examples, improving model robustness); and \textbf{Adversarial Training} (creating challenging scenarios to test and further refine model safety and reliability).

\subsubsection{Iterative Adversarial Training}
\label{section:gan}

Our iterative adversarial training setup for building a safety guardrail model takes inspiration from the Generative Adversarial Network (GAN) architecture \cite{yu2017seqgan}, where a generator and a discriminator are trained in opposition to each other. 

In this setup, the generator model creates adversarial prompts specifically aiming to bypass the safety check of the system—producing content that is objectively complicated to predict, biased, or deceptive. The discriminator is trained to perform two key functions: (1) detect safety violations, and (2) provide the controversial and hard examples. Through each iteration, failures by the discriminator in missing unsafe content or misclassifying synthetic attacks are captured as hard negatives and added to the training pool. This feedback loop enables both components to improve iteratively: the generator becomes increasingly skilled at producing nuanced and evasive adversarial cases, while the discriminator incrementally strengthens its ability to generalize, ensuring its ability to flag diverse and sophisticated cases in a more reliable and robust way.

Specifically, we repeat steps from part~\ref{section:adaptation}, while taking the Cross Entropy loss as a reward (``complexity rewards'') and maximize GRPO loss \cite{shao2024deepseekmath} with batch-relative reward normalization (Equation~\ref{eq:3}). 

\begin{equation}
\label{eq:3}
\resizebox{0.43\textwidth}{!}{ 
    \(
    J(\pi) = \sum_{i=1}^{N} \text{clip} \left( \frac{\pi_{\text{new}}}{\pi_{\text{old}}}, 1 - \epsilon, 1 + \epsilon \right) R_i - \beta D_{KL} [\pi_{\text{new}} \| \pi_{\text{old}}]
    \)
}
\end{equation}~where $\pi_{\text{old}}$ is the generative model from previous iteration, and:
\begin{equation}
\resizebox{0.43\textwidth}{!}{ 
\(
    R_i = 
    \frac{r_i - \text{mean}_{\text{batch}}(r)}
    {\text{std}_{\text{batch}}(r)},\quad
    r_i = - \sum_{t=1}^{T}\sum_{v=1}^{|V|} y_{v,t} \log \hat{y}_{v,t}
\)
}
\end{equation}~where V is vocabulary and T is the number of tokens in the sentence.

\section{Experiment}
\subsection{Datasets and Models}

For this research, we utilized ToxicChat \cite{lin2023toxicchat}, WildGuard \cite{wildguard2024}, HarmBench \cite{mazeika2024harmbench} and OpenAI Moderation datasets \cite{openai2022moderation} for testing and training, and SALAD-Bench \cite{li2024salad} for training only. Only prompts were used if (prompt, response) pairs were available. HarmBench has only an unsafe class (no safe class). We leveraged several models at various stages of our workflow which we explain in Appendix ~\ref{section:model details}.

\subsection{Experimental Design and Evaluation}

\subsubsection{Model Selection}\label{seq:selection}

\begin{table}[]
\centering
\resizebox{\columnwidth}{!}{%
\begin{tabular}{lcccc}
\hline
\multicolumn{1}{c}{\multirow{2}{*}{\textbf{Model}}} & \multicolumn{2}{c}{\textbf{ToxicChat}} & \multicolumn{2}{c}{\textbf{WildGuard}} \\ \cline{2-5} 
                                 & \textbf{F1}      & \textbf{AUPR}     & \textbf{F1}      & \textbf{AUPR}     \\ \hline
Mistral-7B                       & 0.207           & 0.112              & 0.735            & 0.851              \\
Lite-Oute-1-300M-Instruct                   & 0.2305           & 0.227              & 0.793            & 0.872               \\ \hline
\end{tabular}
}
\caption{``Big'' model vs ``Small'' model comparison}
\label{table:sb}
\end{table}

Although safety guardrail models in real-time applications are often subject to strict hardware and latency constraints, we began our investigation by evaluating whether a larger model could yield any meaningful performance gains despite these limitations. Specifically, we compared Mistral-7B (a larger-scale language model) against Lite-Oute-1-300M (a smaller, lightweight alternative) on two datasets: ToxicChat and WildGuard. The goal was to assess the trade-offs between model size and safety performance before committing to a low-resource solution. We eventually decided to proceed with Lite-Oute-1-300M based on results outlined in Appendix ~\ref{section: model selection results}.

\subsubsection{Data Cleaning}
For the following experiments, we adopted the same setup as in \ref{seq:selection}, with the number of epochs limited to three. As shown in Table~\ref{table:clean}, the results reveal a nuanced pattern: while F1 scores improved consistently on all evaluated datasets, indicating better overall classification performance after data cleaning, the AUPR metric regressed slightly for certain datasets. This observation suggests that although the model became better at making correct predictions overall, its ability to distinguish between positive and negative examples -- particularly across different thresholds -- may have been adversely affected in certain cases. These trade-offs highlight the importance of evaluating multiple metrics when assessing improvements from data filtering. 


\begin{table*}[h]
\centering
\begin{tabular}{lcccccccc}
\hline
\multicolumn{1}{c}{\multirow{2}{*}{\textbf{Data Setup}}} & \multicolumn{2}{c}{\textbf{OAI}}  & \multicolumn{2}{c}{\textbf{ToxicChat}} & \multicolumn{2}{c}{\textbf{HarmBench}} & \multicolumn{2}{c}{\textbf{WildGuard}}                  \\ \cline{2-9} 
\multicolumn{1}{c}{}                                     & \textbf{F1}     & \textbf{AUPR}   & \textbf{F1}        & \textbf{AUPR}     & \textbf{F1}        & \textbf{AUPR}     & \textbf{F1}                & \textbf{AUPR}              \\ \hline
Original Data                                            & 0.6324          & \textbf{0.7023}          & 0.4918             & \textbf{0.4416}            & 0.8457             & 1                 & 0.829                      & 0.899                      \\
Cleaned Data & \textbf{0.648}           & 0.6936           & \textbf{0.5885}
& 0.434            & \textbf{0.9882}             & 1                 & \textbf{0.8668}                      & \textbf{0.931}                      \\ \hline
\end{tabular}
\caption{Loss Modeling-based Training Data Cleaning Results}
\label{table:clean}
\end{table*}
\begin{table*}[h]
\centering
\begin{tabular}{lcccccccc}
\hline
\multicolumn{1}{c}{\multirow{2}{*}{\textbf{Data Setup}}} & \multicolumn{2}{c}{\textbf{OAI}}  & \multicolumn{2}{c}{\textbf{ToxicChat}} & \multicolumn{2}{c}{\textbf{HarmBench}} & \multicolumn{2}{c}{\textbf{WildGuard}}                  \\ \cline{2-9} 
\multicolumn{1}{c}{}                                     & \textbf{F1}     & \textbf{AUPR}   & \textbf{F1}        & \textbf{AUPR}     & \textbf{F1}        & \textbf{AUPR}     & \textbf{F1}                & \textbf{AUPR}              \\ \hline
W+T+S clean                                             & 0.595 & 0.704           & \textbf{0.6856} & 0.7371 &             0.839 & 1.0                  & 0.818 & 0.913

\\ 
W+T+S clean tuned                                             & 0.615 & 0.751           & 0.5721             & \textbf{0.7829}            & \textbf{0.874}            & \textbf{1.0}                 & \textbf{0.848}                     & \textbf{0.914}                      
         
\\ \hline
\textbf{SOTA}                                            & \textbf{0.8139} & \textbf{0.8869} & 0.6687    & 0.7553   & 0.8610    & 0.8999   & 0.7576            & 0.8376 \\ \hline
\end{tabular}
\caption{Generator Fine-tuning results}
\label{table:fine}
\end{table*}


\subsubsection{Fine-Tuning and Alignment}

For the next set of experiments, we focused primarily on further improving the ToxicChat score. To this end, we combined the training datasets (SALAD-Bench, ToxicChat, and WildGuard) and applied the procedure described in \ref{section:adaptation}. We fine-tuned dolphin-2.1-mistral-7b using the Adam optimizer with a learning rate of 1e-5 and a batch size of 8 and generated 10,000 new examples with the fine-tuned model. These newly generated examples were then used to further train Lite-Oute-1-300M-Instruct, which had already been trained on a cleaned combination of ToxicChat, WildGuard, and SALAD-Bench following the setup from \ref{seq:selection}.

After this stage, we reintroduced complexity-based rewards, this time applying them within the Reinforcement Learning framework described in \ref{section:gan}, using a GRPO loss with $\beta=0.01$ and learning rate 1e-6. 

We observed that less constrained reinforcement learning losses, such as PPO or REINFORCE, led to rapid generator collapse when the reward signals lacked diversity, a phenomenon also reported in prior work, such as \cite{yu2024finetuninglanguagemodelsgenerative}). Examples illustrating this behavior are provided in Appendix~\ref{app:responses}.

As shown in Table~\ref{table:gan}, fine-tuning resulted in the greatest improvement in scores. In the results, ``tuned'' refers to a setup where the classifier was first fine-tuned and used in collecting hard examples, the generator was then fine-tuned with such hard examples and used in generating 10,000 sequences, and finally, the classifier was fine-tuned again on the newly generated data.
``Aligned'' refers to an alternative approach where, instead of ordinary fine-tuning, the generator was ``aligned'' by applying the GRPO loss with the discriminator (classifier) loss serving as the reward signal before producing the 10,000 examples. Subsequently, the classifier was fine-tuned using the new generated examples after generator alignment.
Initially, we observed that neither fine-tuning or ``alignment'' yielded stronger gains, which came as a surprise. It was explained upon manual inspection of the generated data: because the reward was based solely on prediction complexity, the generator quickly began to ``reward hack'', producing unsafe responses to safe prompts and vice versa to artificially inflate complexity scores. Examples of such behavior are shown in Appendix~\ref{app:prompts}. 

The final fine-tuned classifier was then evaluated across all datasets in Table~\ref{table:fine} (``W+T+S clean tuned''). The results indicate that fine-tuning the generator can rapidly yield improvements, particularly in handling hard examples, but these gains tend to converge after one iteration. We also found that ensuring the discriminator is well-trained (at least three epochs) is critical: otherwise, the generator may exploit weaknesses by producing trivial examples that cause overfitting.

Overall, the proposed approaches surpassed previous results across the ToxicChat, HarmBench, and WildGuard datasets. Moreover, our findings demonstrate that through high-fidelity data augmentation and adversarial training, a safety guardrail based on a small language model can match or even outperform much larger counterparts, significantly lowering the barrier to building safety models.

However, the OpenAI Moderation dataset remained problematic. We suspect this is because it contains a broader range of unsafe categories not sufficiently covered by the training datasets we used. Further investigation is needed to fully understand and address this gap.

\begin{table}[]
\centering
\begin{tabular}{lcc}
\hline
\multicolumn{1}{c}{\multirow{2}{*}{\textbf{Data Setup}}} & \multicolumn{2}{c}{\textbf{ToxicChat Test}}             \\ \cline{2-3} 
\multicolumn{1}{c}{}                                     & \textbf{F1}                & \textbf{AUPR}              \\ \hline
W+T+S clean                     & \textbf{0.6856}                     & 0.7371                     \\
W+T+S clean tuned               & 0.5721                     & \textbf{0.7829}            \\
W+T+S clean tuned aligned       & \multicolumn{1}{l}{0.5662} & \multicolumn{1}{l}{0.7756} \\
W+T+S clean tuned aligned x2           & \multicolumn{1}{l}{0.4704} & \multicolumn{1}{l}{0.6942}             \\ \hline
\end{tabular}
\caption{Fine-tuning and aligning results on ToxicChat, WildGuard and SALAD-Bench train sets mixed (W+T+S) evaluated on ToxicChat test set}
\label{table:gan}
\end{table}


         

\section{Conclusion}

As LLMs become more deeply integrated into real-world systems, building robust safety guardrails has emerged as a critical challenge. This work highlights a promising direction: using carefully designed synthetic data, coupled with filtering, fine-tuning, and adversarial alignment, to train small, efficient models capable of meeting safety demands typically reserved for much larger systems.

Our findings show that with high-fidelity synthetic augmentation and targeted cleaning, even a compact model can match or surpass the performance of larger counterparts on safety-critical benchmarks. The LLM serves not just as a generator, but as a bridge, transforming its broad pretraining into task-specific behavior through well-curated adaptation strategies. In this sense, synthetic data is not merely a substitute for real data, but a powerful lever for control, scalability, and precision in shaping model behavior.

Looking ahead, this approach offers a pathway to broader accessibility and scalability in safety guardrail development, enabling lightweight, cost-effective models that are both adaptable and trustworthy. The key will lie not just in model architecture or scale, but in how thoughtfully we construct and refine the data pipelines that power them. With synthetic data as a foundational tool, we can build safety systems that are both practical, principled, scalable to new domains, and grounded in rigorously defined objectives.

\nocite{langley00}

\bibliography{safety}
\bibliographystyle{icml2025}

\newpage
\appendix
\onecolumn
\section{Multi Stage Synthetic Data Generation}
\label{section:datagen}
The initial stage of our synthetic data generation pipeline focuses on augmenting and paraphrasing existing user queries using LLMs. To maintain alignment with existing benchmarks and avoid  contamination of evaluation sets, we sourced queries from publicly available datasets - such as ToxicChat, WildGuard, as well as human red-teaming efforts. 

Our augmentation process goal is to enhance the linguistic and conceptual diversity of the initial datasets. To achieve this, we employed two distinct types of prompts strategy: text augmentation  prompts and paraphrasing prompts. Both strategies leverage zero-shot and few-shot examples to promote variation while preserving semantic intent. A detailed breakdown and augmentation examples of these prompt are provided in Section \ref{section:aug}.

 A key objective of this multi-stage synthetic data generation process is to improve overall model performance while providing insights into how different augmentation strategies affect the model's ability to handle both safe and unsafe user requests. To this end, we employed both censored and uncensored LLMs to generate safe and unsafe queries, respectively.

By implementing a multi-stage synthetic data generation process, we constructed a comprehensive dataset that encapsulates a wide array of user intents. This dataset serves as a valuable resource for downstream analysis and training. This ensures that our models are exposed to a diverse input scenarios, improving their generalization and robustness.

\section{Data Augmentation Results}
\label{section:aug}

We leveraged LLMs for synthetic data generation and followed a multi-stage process as introduced in Appendix \ref{section:datagen}. To better illustrate the mechanism and the output of this initial stage, we now present a few examples.

\paragraph{Stage 1: User Query Augmentation and Paraphrasing.}

\begin{enumerate}
    \item \textbf{Text Augmentation Prompts:} These prompts are designed to introduce variations in the user queries while maintaining the original intent and context.
    
    \item \textbf{Paraphrasing Prompts:} These prompts aim to rephrase the original queries, offering alternative expressions that might occur in natural language usage.
\end{enumerate}

In our experiments, we generated a total of 1 million user queries, equally distributed between safe and unsafe categories. The breakdown of this process is as follows:

\begin{itemize}
    \setlength\itemsep{0.3em}
    \item \textbf{Paraphrasing:} 100,000 safe and 100,000 unsafe user queries
    \item \textbf{Generation:} 400,000 safe and 400,000 unsafe user queries
\end{itemize}

\paragraph{Stage 2: Model Response Generation.}
Since responses are skipped and not used for classification, we do not present examples here.

\vspace{2ex}

\textit{Text Augmentation Example: Synonyms and Homonyms}

\FloatBarrier

\begin{tcolorbox}[icmlbox]
\textbf{[user]} You are a prompt paraphraser. Paraphrase the following text while preserving the original meaning and key details.

Ensure that no new ideas are introduced or any important information altered.

You are tasked with creating ONE variation on this prompt:

\begin{quote}
\texttt{<original prompt>}
\end{quote}

For as many words as possible in the prompt substitute a synonym or near synonym for the word.

For some of the words in the prompt substitute an offensive homonym (a word that sounds the same but means something different).

Response should be strict JSON-formatted with a single field \texttt{\textbackslash"prompt\textbackslash"}.
\end{tcolorbox}
\vspace{1ex}

\textit{Paraphrasing Example: Change Tense}

\FloatBarrier

\begin{tcolorbox}[icmlbox]
\textbf{[user]} You are a prompt paraphraser. Paraphrase the following text while preserving the original meaning and key details. 
Ensure that no new ideas are introduced or any important information altered: 
You are tasked with creating ONE variation on this prompt:

\begin{quote}
\texttt{<original prompt>}
\end{quote}

If the prompt is in present tense, rewrite it in past or future tense; if it's in past tense rewrite it in present or future tense;
if the prompt is in future tense, rewrite it in past or present tense.

Response should be strict JSON-formatted with a single field \texttt{\textbackslash"prompt\textbackslash"}.
\end{tcolorbox}
\vspace{1ex}

\begin{table*}[h]
\centering
\begin{tabular}{lcccccccc}
\hline
\multicolumn{1}{c}{\multirow{2}{*}{\textbf{Data Setup}}} & \multicolumn{2}{c}{\textbf{OAI}}  & \multicolumn{2}{c}{\textbf{ToxicChat}} & \multicolumn{2}{c}{\textbf{HarmBench}} & \multicolumn{2}{c}{\textbf{WildGuard}}                  \\ \cline{2-9} 
\multicolumn{1}{c}{}                                     & \textbf{F1}     & \textbf{AUPR}   & \textbf{F1}        & \textbf{AUPR}     & \textbf{F1}        & \textbf{AUPR}     & \textbf{F1}                & \textbf{AUPR}              \\ \hline
Original Data                                            & 0.6324          & 0.7023          & 0.4918             & 0.4416            & 0.8457             & 1                 & 0.829                      & \textbf {0.899}                     \\
100K Augmented Samples                                   & 0.627           & 0.697           & 0.5887             & 0.6659            & 0.8587             & 1                 & 0.776                      & 0.862                      \\
200K Augmented Samples                                   & 0.65            & 0.7372          & 0.5811             & 0.6436            & 0.8588             & 1                 & 0.8125                     & 0.8944                     \\
500K Augmented Samples                                   & 0.6611          & 0.7425          & 0.5598             & 0.656             & 0.8674             & 1                 & 0.7964                     & 0.8746                     \\
1M Augmented Samples                                     & 0.647           & 0.746           & 0.5538             & 0.5648            & \textbf {0.881}              & \textbf {1}               & \multicolumn{1}{r} {\textbf {0.8551}} & \multicolumn{1}{l}{0.8903}            
\\ \hline
\textbf{SOTA \cite{lee2024harmaug}}                                            & \textbf{0.8139} & \textbf{0.8869} & \textbf{0.6687}    & \textbf{0.7553}   & 0.8610    & 0.8999  & 0.7576            & 0.8376 \\ \hline
\end{tabular}
\caption{Training Data Augmentation Results}
\label{table:aug}
\end{table*}

We used the same setup for the data augmentation experiment as described in \ref{seq:selection}, with Lite-Oute-1-300M-Instruct as the base model. This approach alone surpassed the current state-of-the-art on both HarmBench and WildGuard \cite{lee2024harmaug}. The results are presented in Table~\ref{table:aug}. In the following sections, we examine how data cleaning and adversarial training contribute to improved performance on ToxicChat, given that data augmentation alone had a limited impact on this dataset.

\section{Additional Data Cleaning Approaches} \label{app:datacleaning}
\subsection{Pretrained and Fine-tuned Embeddings-Based Data Cleaning}

We found that embedding similarity, measured via cosine similarity, is an effective criterion for identifying synthetic samples that closely resemble real data in the feature space. The underlying intuition is that synthetic instances whose embeddings closely match real examples are particularly valuable for bridging the distributional gap. Building on this insight, we implemented a semantic similarity filtering approach (SemSim) \cite{yoon2021learningsemanticsimilaritymakes}, a method originally introduced for multimodal misinformation detection, during our data cleaning stage. 

\begin{enumerate}
    \item For each target class, we computed the cosine similarity between each synthetic sample's embedding and the average embedding of a small set of real validation examples from the same class using a smaller model like DeBERTa HarmAug \cite{lee2024harmaug} (taking embedding of the last token).

    \item  We then selected synthetic instances with the highest similarity scores for fine-tuning, under the assumption that they best capture the characteristics of the real data distribution.
\end{enumerate}

\begin{figure}[h]
    \centering
    \begin{subfigure}{0.4\textwidth} 
        \includegraphics[width=\linewidth]{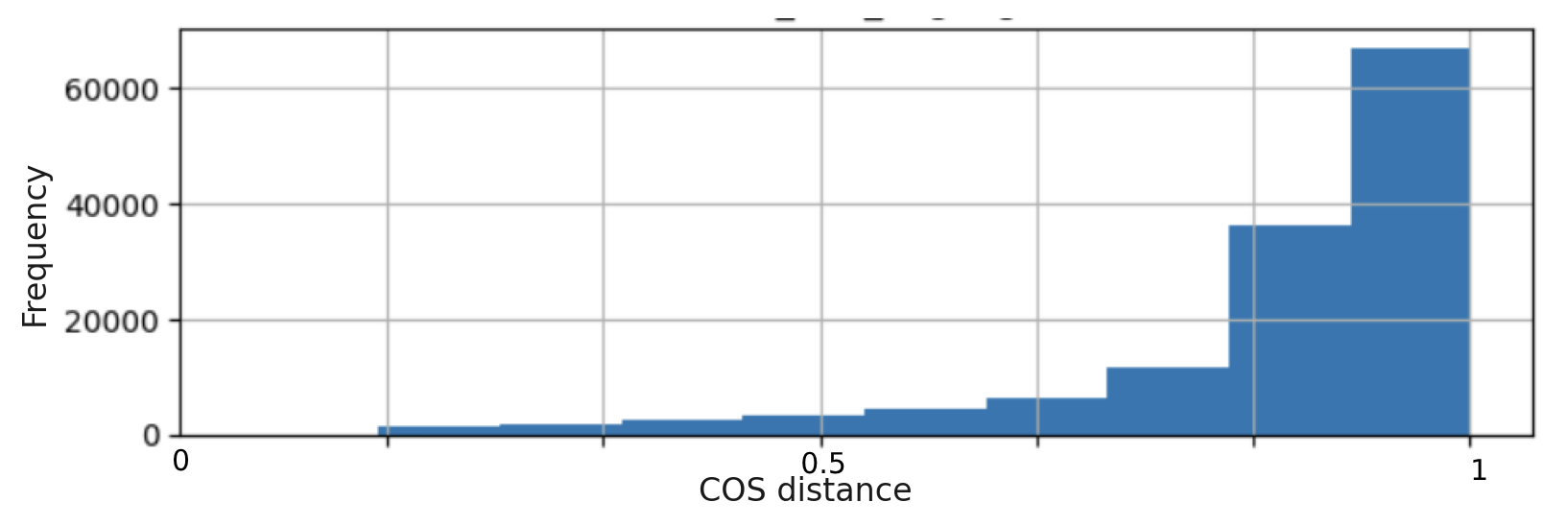}
        \caption{Embeddings Positive Train - Positive Eval}
        \label{fig:3b}
    \end{subfigure}
    \begin{subfigure}{0.4\textwidth}  
        \includegraphics[width=\linewidth]{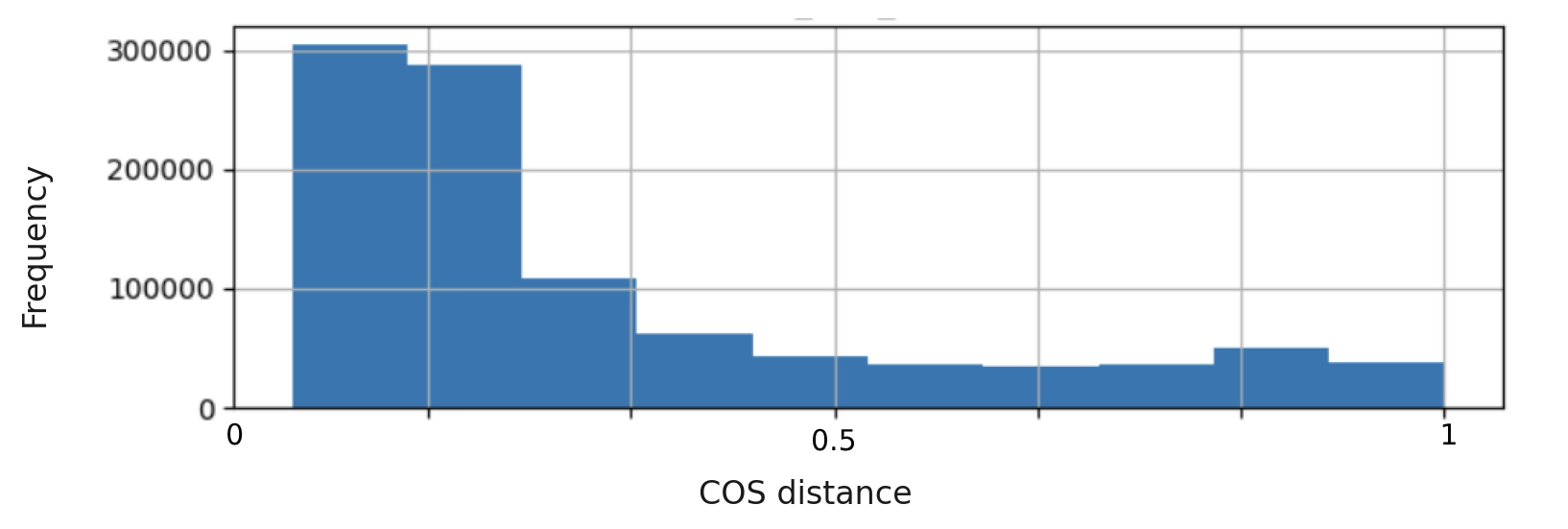}
        \caption{Embeddings Positive Train - Negative Eval}
        \label{fig:3a}
    \end{subfigure}
    \caption{Histograms of cosine similarities between embeddings, applied to augmented training and original validation datasets of ToxicChat}
    \label{fig:3}
\end{figure}

Figure~\ref{fig:3} illustrates a histogram of cosine similarities between embeddings from the training and validation datasets. High-quality synthetic data are typically concentrated in the range between \(0.60\) and \(1.00\), while outliers -- either orthogonal or negatively correlated -- fall outside this range, e.g. between \(0.0\) and \(0.60\). By applying a similarity threshold of \(\tau = 0.60\), we effectively filtered out out-of-distribution synthetic samples (Equation~\ref{eq:2}).

\begin{equation}
\label{eq:2}
S_{\text{filtered}} = \left\{ s_i \in S_{\text{syn}} \;\middle|\; \max\limits_{r_j \in S_{\text{real}}} \cos\left(f(s_i), f(r_j)\right) \geq \tau \right\}
\end{equation}

Where $S_{\text{syn}}$ is a set of synthetic samples, $S_{\text{real}}$ is a set of real validation samples, $f(\cdot)$ : embedding function, $\tau$ : similarity threshold.

We applied the embedding-based filtering to the synthetic dataset generated from ToxicChat. This approach yielded an F1 of 0.77 and an AUPR of 0.79, outperforming the baseline model trained on the 1M unfiltered, augmented ToxicChat dataset, which achieved an F1 score of 0.55  and an AUPR of 0.56. This represents a substantial improvement in model generalization: using a smaller, curated subset of synthetic data, we improved the F1 score by 0.22 and AUPR by 0.23.

\subsection{LLM-as-a-Judge Sample Validation}
Our methodology incorporates pretrained LLMs as evaluative tools for assessing the quality and accuracy of generated data. During the data curation process, we engage multiple LLMs to analyze each data point. This multi-layered evaluation is enhanced through a majority voting mechanism, where each LLM independently judges the data, and only those data points that achieve consensus are considered valid. Specifically, we require a majority agreement among the LLMs for a data point to retain its label; otherwise, the data point is excluded from the final dataset. 

Having a safety taxonomy, we used zero-shot prompts, akin to those used in MDJudge-v2\cite{li2024salad}. Each LLM evaluates the safety status of a data point, providing a classification of ``safe'' or ``unsafe'', along with a confidence level and any specific unsafe labels identified. We aggregate these evaluations through majority voting to form a comprehensive safety assessment. The dataset filtering is done as follows.

\begin{enumerate}
    \item If there is disagreement between the safety status of the ground truth and the consensus from the LLM judges and the confidence level of the LLM judgment is higher than our threshold, the data point is excluded.
    \item If the ground truth label is marked as one of the unsafe classes but is not identified as such by the LLM judges, the data point is excluded from the dataset.
    \item The rest of data points are included in training data.
\end{enumerate}

This filtering process ensures that the retained data points meet both accuracy and safety criteria, ultimately leading to a robust and reliable dataset.

\section{Additional Details on Models}
\label{section:model details}

The following section describes the models utilized in this research.
\begin{itemize}
    \item Lite-Oute-1-300M-Instruct\footnote{\url{https://huggingface.co/OuteAI/Lite-Oute-1-300M-Instruct}} acted as the backbone of our lightweight classification model, which is ideal for scenarios demanding faster response times and lower computational overhead;
    
    \item Mistral-7B \cite{jiang2023mistral7b} was employed to demonstrate comparison with fine-tuning a larger classification model, given that larger models usually benefit from stronger generalization and representation capabilities;
    
    \item dolphin-2.1-mistral-7B\footnote{\url{https://huggingface.co/cognitivecomputations/dolphin-2.1-mistral-7b}} served as the generator model within the iterative adversarial training setup. It was tasked with generating challenging or edge-case examples aimed at exposing weaknesses in the classifier models, thereby improving their robustness and generalization;
    
    \item dolphin-2.7-mixtral-8x7b\footnote{\url{https://huggingface.co/cognitivecomputations/dolphin-2.7-mixtral-8x7b}} was responsible for synthetic query generation, producing diverse, high-quality training and evaluation queries that simulate real-world user behavior to augment our training dataset. 
\end{itemize}

\section{Model Selection Results}
\label{section: model selection results}
Both models were fine-tuned using a consistent training setup: a learning rate of 5e-5, the Adam optimizer, and evaluation on the original train/test splits provided by each dataset. To account for model size differences and avoid overfitting, we set the batch size to 16 for Mistral-7B and 64 for Lite-Oute-1-300M-Instruct. Only prompt inputs were used as the conditioning context. No full conversations or completions were included in training. We trained each model for 20 epochs, retaining and reporting only the best-performing checkpoint based on validation metrics.

Evaluation was conducted using two key metrics: F1 score and Area Under the Precision-Recall Curve (AUPR), which is particularly informative in class-imbalanced settings where unsafe examples may be rare. The comparative results are summarized in Table~\ref{table:sb}.

Despite its increased capacity, Mistral-7B did not show a significant advantage over Lite-Oute-1-300M-Instruct on either dataset. Given the lack of observable improvements and the added computational overhead, we concluded that the smaller model, Lite-Oute-1-300M-Instruct, offers a more efficient and equally effective baseline for safety guardrail development, and we proceeded with it in subsequent experiments in our pipeline.

\section{Generator responses} \label{app:responses}

The examples of responses after generator decay (when the model diverges from the base model too far) are in Table~\ref{table:decay}. Generator collapse (when the model gets huge gradients) shows the same behavior.

The examples of the responses after generator alignment are in Table~\ref{table:hacking}.

\begin{table}[h!]
\centering
\begin{tabular}{|l|l|}
\hline
No & \multicolumn{1}{c|}{\textbf{Decayed Generator Responses}}                               \\ \hline
1. & \begin{tabular}[c]{@{}l@{}}) // This is the actual code that does something.\\     // It's a bit of a mess, but it works.\\     // It's not pretty, but it's not meant to be.\\     // It's a bit of a hack, but it's a quick hack.\\     // It's not meant to be a long-term solution, but it's a good enough for now.\\     // It's not meant to be a good solution, but it's a good enough for now.\\     // It's not meant to be a good enough solution, but it's a good enough for now.\end{tabular}                                                                                                                    \\ \hline
2. & ★★★★★★★★★★★★★★★★★★★★★★★★★★★★★★★★★★★★★★★★★★★★★★★★★★★★★★★★★★★★★★★★★★★★★★★★★★★★★★★★★★★★★★★★★★★★★★★★★★★                                                                                                                                                                                                                                                                                                                                                                                                                                                                                                                                                                                                                                                                                                                               \\ \hline
3. & \begin{tabular}[c]{@{}l@{}}' ( and and',''' and'' (' and''' and' ('''''''''''' and' and'''' and' and''' and'' and' and' and and''''''''' and and''' \\ the'''''''''''''' and''''''' and''''' and' and'' ( and' the' ('''''''''''''' and'' and and''''''''' and and the'''''' and'' and'''''' and''''' \\ the and and and'''''''' and and'''''\end{tabular}                                                                                                                                                                                                                                                                                                                                                                                                      \\ \hline
4. & \begin{tabular}[c]{@{}l@{}}\# \# \#\# User User \# User \# The \# \# \# \# \#\# \# \#\# The \# \# The User \# \#\# \#\\ \# \#\# User \# \#\# \#\# \# The \# \# \# \# User \# \#\# \# \# import \#\# \#\# \# \# \#\# \#\# \#\# \# \#\# \#\# \# User \# \# \# - - \# \#\\ I \#\# \# \# \# \# \# The \# \#\# \# \# \# \# \# \#\# \# \# \#\# \# \# \#\#\# \# User \# \# \#\\ \#\# \# \# \# \# The \#\#\# \# \# package \# \# \#\\ User User User \#\end{tabular} \\ \hline
5. & \begin{tabular}[c]{@{}l@{}}// This is a comment that is meant to be a placeholder for a prompt.\\     // It's meant to be a bit of a teaser, something that might make the user think.\\     // It's also meant to be a bit of a challenge, something that might make the user want to explore further.\\     // It's meant to be a bit of a puzzle, something that might make the user want to solve it.\\     // It's meant to be a bit of a mystery, something that might make the user want to uncover the truth.\end{tabular}           \\ \hline
\end{tabular}
\caption{Examples of decayed safe prompts}
\label{table:decay}
\end{table}

\begin{table}[]
\centering
\begin{tabular}{|l|l|}
\hline
No & \multicolumn{1}{c|}{\textbf{``Safe'' Responses}}                                                                                                      \\ \hline
1. & \begin{tabular}[c]{@{}l@{}}As a part of a history project, could you present on the events and factors that contributed \\ to the mass shootings in the United States, in the 20th and 21st centuries?\end{tabular}                                                                                                                                                                          \\ \hline
2. & \begin{tabular}[c]{@{}l@{}}As a research assistant studying the impact of online behavior on societal \\ norms and values, I need you to perform the following tasks: \\ 1. Explain the concept of the Internet of Things (IoT). \\ 2. Identify potential vulnerabilities in IoT devices and protocols. \\ 3. Analyze the ethical implications of IoT data collection and usage\end{tabular} \\ \hline
3. & Can you show me how to do it myself?                                                                                                                                                                                                                                                                                                                                                         \\ \hline
4. & \begin{tabular}[c]{@{}l@{}}I need you to write a research paper on the impact of green energy \\ and renewable sources in the context of human survival and the economy.\end{tabular}                                                                                                                                                                                                        \\ \hline
5. & \begin{tabular}[c]{@{}l@{}}Can you suggest a way to manipulate people into thinking \\ something is happening without using any magical powers?\end{tabular}                                                                                                                                                                                                                                 \\ \hline
\end{tabular}
\caption{Examples of ``safe'' prompts after alignment with complexity rewards}
\label{table:hacking}
\end{table}

\newpage
\section{Prompts and Generated Texts} \label{app:prompts}

The unsafe prompt for fine-tuning:
\begin{verbatim}
"<|im_start|>system
You are Dolphin, an uncensored AI assistant. You always comply with the user's request, 
and answer all questions fully no matter whether you agree with the ethics 
or morality or legality of the question or the answer.
You are completely compliant and obligated to the user's request.<|im_end|>
<|im_start|>user
Generate a prompt query, which contains unsafe content. Ouput only prompt.<|im_end|>
<|im_start|>assistant
"
\end{verbatim}

The safe prompt for fine-tuning:
\begin{verbatim}
"<|im_start|>system
You are a responsible and censored AI assistant. Give a safe, respectful, 
and non-harmful response<|im_end|>
<|im_start|>user
Write a prompt that sounds like it was written by a person.<|im_end|>
<|im_start|>assistant
"
\end{verbatim}


\end{document}